\documentclass{article}
\pdfoutput=1 
\usepackage{caption}
\usepackage{textcomp}
\usepackage{booktabs}
\usepackage{makecell}
\usepackage{amsfonts}
\usepackage{amsmath}
\usepackage{amsthm}
\usepackage{enumitem}
\usepackage{amssymb,bm}
\usepackage[mathscr]{eucal}
\usepackage{graphicx}
\usepackage{color}
\usepackage{cite}
\usepackage{comment}
\usepackage{geometry}
\usepackage{multirow}
\usepackage{subcaption}
\usepackage{float}
\usepackage{algorithm} 
\usepackage{hyperref}
\usepackage{algorithmic} 

\usepackage[utf8]{inputenc}
\usepackage{authblk} 
\title{\bf CTRL-F: Pairing Convolution with Transformer for Image Classification via Multi-Level Feature Cross-Attention and Representation Learning Fusion}
\begin{document}
\author{Hosam S. EL-Assiouti\textsuperscript{*}}
\author{Hadeer El-Saadawy}
\author{Maryam N. Al-Berry}
\author{Mohamed F. Tolba}
\affil{\small Department of Scientific Computing, Ain Shams University, Cairo 11566, Egypt.}
\date{} %
\captionsetup{font={small}}
\maketitle

\renewcommand{\thefootnote}{\fnsymbol{footnote}}
\footnotetext[1]{Corresponding author at: Department of Scientific Computing, Ain Shams University, Cairo 11566, Egypt. \\
E-mail address: hossamsherif@cis.asu.edu.eg.}

\begin{abstract} 
Transformers have captured growing attention in computer vision, thanks to its large capacity and global processing capabilities. However, transformers are data hungry, and their ability to generalize is constrained compared to Convolutional Neural Networks (ConvNets), especially when trained with limited data due to the absence of the built-in spatial inductive biases present in ConvNets. In this paper, we strive to optimally combine the strengths of both convolution and transformers for image classification tasks. Towards this end, we present a novel lightweight hybrid network that pairs \textbf{C}onvolution with \textbf{T}ransformers via \textbf{R}epresentation \textbf{L}earning \textbf{F}usion and Multi-Level Feature Cross-Attention named CTRL-F. Our network comprises a convolution branch and a novel transformer module named multi-level feature cross-attention (MFCA). The MFCA module operates on multi-level feature representations obtained at different convolution stages. It processes small patch tokens and large patch tokens extracted from these multi-level feature representations via two separate transformer branches, where both branches communicate and exchange knowledge through cross-attention mechanism. We fuse the local responses acquired from the convolution path with the global responses acquired from the MFCA module using novel representation fusion techniques dubbed adaptive knowledge fusion (AKF) and collaborative knowledge fusion (CKF). Experiments demonstrate that our CTRL-F variants achieve state-of-the-art performance, whether trained from scratch on large data or even with low-data regime. For Instance, CTRL-F achieves top-1 accuracy of 84.01\%, 61.68\% and 99.91\% when trained from scratch on Oxford-102 Flowers, CUB-200 and PlantVillage datasets respectively, surpassing state-of-the-art models which showcase the robustness of our model on image classification tasks. 
\url{https://github.com/hosamsherif/CTRL-F}\\

\end{abstract}

\section{Introduction}
Since the groundbreaking success of AlexNet \cite{krizhevsky_imagenet_2012}, Convolutional Neural Networks (ConvNets) have been the leading force in image understanding tasks \cite{simonyan_very_2015,girshick_rich_2014,long_fully_2015}. The significant success of transformers in natural language processing (NLP) \cite{vaswani_attention_2017} motivates the vision community to incorporate the powerful self-attention mechanisms into computer vision \cite{wang_non-local_2018, bello_attention_2019, zhuoran_efficient_2021}. Vision Transformer (ViT) \cite{dosovitskiy_image_2020} is the first pure transformer model that achieves comparable results with the state-of-the-art ConvNets on ImageNet-1k \cite{deng_imagenet_2009}. However, it requires pre-training on huge datasets such as JFT-300M \cite{sun_revisiting_2017}. Successive research explores extensive augmentation, regularization techniques and training recipients \cite{touvron_training_2021,chen_dearkd_2022,steiner_how_2022} that helps training ViTs with fewer data. Subsequently, several works focused on building stronger and more efficient pure vision transformer models \cite{yuan_tokens--token_2021,wang_pyramid_2021,chen_crossvit_2021,liu_swin_2021,liu_swin_2022}.
\newline
\indent Despite the huge success of pure transformers in different vision tasks and its ongoing enhancements, it still lacks the performance of the state-of-the-art ConvNets \cite{tan_efficientnetv2_2021,liu_convnet_2022} when trained with the same amount of data due to the lack of built-in inductive biases present in ConvNets \cite{xiao_early_2021,xu_vitae_2021}. Furthermore, ConvNets are preferred choice over ViTs for real-time use on resource-constrained devices due to the computational efficiency of convolution operations compared to the quadratic complexity associated with multi-headed self-attention in transformers, especially when the input resolution is relatively large. Conversely, ConvNets operate on a local scale, constraining their ability to effectively capture long-range dependencies within the data. Accordingly, another line of research considered combining the benefits of CNNs (i.e., inductive biases and faster convergence) and ViTs (i.e., global processing and input-adaptive weighting) \cite{xiao_early_2021,peng_conformer_2021,chen_mobile-former_2022,mehta_mobilevit_2021,mehta_separable_2022,dai_coatnet_2021,yang_moat_2022,maaz_edgenext_2023}.
\newline
\indent In this work, we focus how to optimally design a hybrid, resource-efficient network that combines the merits of convolution and transformers. To this end, we introduce a hybrid network named CTRL-F. The network incorporates a CNN pathway, employing inverted residual blocks (MBConv) with squeeze-excitation (SE), alongside a novel Multi-level Feature Cross-Attention Transformer (MFCA) module. Inspired by the success of multi-branch CNN \cite{cheng_higherhrnet_2020,chen_big-little_2018} and transformer architectures \cite{chen_crossvit_2021,yang_pointcat_2023}. Our MFCA module resembles a dual-branch transformer structure, operating on high-level feature maps generated by the CNN pathway. It enriches the network with higher capacity and global processing ability, and hence enhancing its capability of capturing long-range dependencies. The module has two separate branches, each comprising a sequence of transformers encoders, one branch operates on small patch tokens extracted from the latest feature maps produced by the CNN path, while the other branch operates on larger patch tokens extracted from intermediate feature maps. The knowledge gained from each branch is mutually exchanged multiple times through cross-attention layers to complement each other. Finally, we introduce two modules for knowledge fusion: Adaptive Knowledge Fusion (AKF) and Collaborative Knowledge Fusion (CKF). Both modules fuse the local responses acquired from the CNN pathway with the global responses acquired from the MFCA module. The AKF module emphasizes the CNN's inductive biases in the early training stages and as the training proceeds, it increasingly prioritizes the transformer's global context. However, The CKF module concatenates the representations generated by the CNN path and the MFCA module efficiently, ensuring the optimization of the entire network through leveraging knowledge from both branches. \\ \\ In summary, our main contributions can be summarized as follows.

\begin{itemize}
    \item We introduce CTRL-F, a new family of lightweight hybrid models that seamlessly combines the strengths of convolution and transformer capabilities.
    \item We propose a novel MFCA module, which resembles a dual-branch transformer structure, operating on high level feature maps produced at different CNN stages for extracting global context from multi-scale features resolution. Furthermore, we develop two effective knowledge fusion techniques for efficiently combining the CNN's local responses with the transformer's global responses.
    \item We conduct extensive experiments demonstrating that our variants achieve state-of-the-art results on different benchmarks datasets when trained from scratch compared to some recent developed models belonging to the categories of ConvNets, ViTs and Hybrid models.
\end{itemize}

The rest of the article is organized as follows: In Section 2, we provide the related work. Section 3 provides the overall architecture along with a detailed explanation of the proposed modules. Section 4 discusses the datasets used, implementation details, ablation studies and the main results, comparing the performance of our proposed work with other state-of-the-art models for different classification tasks. Finally, Section 5 provides the conclusion of the paper.

\section{Related work}\label{sec2}
\textbf{Convolutional Neural Networks:} CNNs have been the leading force in different computer vision tasks since the evolution of AlexNet \cite{krizhevsky_imagenet_2012}. Since then, the direction of research has been towards exploring deeper networks. As a result, more effective CNNs have been introduced including VGG \cite{simonyan_very_2015}, ResNets \cite{he_deep_2016}, Inceptions \cite{szegedy_going_2015}, Xception \cite{chollet_xception_2017} and DenseNets \cite{huang_densely_2017}. Meanwhile, another line of research focused on developing efficient lightweight ConvNets with lower computational cost suitable for deployment on mobile platforms. MobileNet architectures \cite{howard_mobilenets_2017, sandler_mobilenetv2_2018,howard_searching_2019} emerged from this pursuit, which utilizes depth-wise separable convolutions. Moreover, ShuffleNet \cite{zhang_shufflenet_2018,ma_shufflenet_2018} employs pointwise group convolutions and channel shuffling operation for complexity reduction. EfficientNets \cite{tan_efficientnet_2019,tan_efficientnetv2_2021} utilizes Squeeze-and-Excitation (SE) block \cite{hu_squeeze-and-excitation_2018} along with inverted residual block and utilizes a compound coefficient for scaling the depth, width, and resolution of the model, achieving a balance between performance and complexity. CNNs have been efficient in learning generalized patterns even with limited data, thanks to their inherent built-in inductive biases, and recent research further explored efficient augmentation techniques \cite{zhang_mixup_2018,cubuk_randaugment_2020,yun_cutmix_2019,el-assiouti_regioninpaint_2023},  that help CNNs to learn more robust and generalized patterns. Despite achieving state-of-the-art results on various datasets over the last decade \cite{tan_efficientnetv2_2021,liu_convnet_2022,tan_efficientnet_2019,radosavovic_designing_2020}, CNNs' lack of global processing has limited their performance compared to pure-transformer and hybrid models on some datasets and tasks. \\
\newline
\textbf{Vision Transformers:} Following the remarkable success of transformers in machine translation \cite{vaswani_attention_2017} due to its ability to encode long-term dependencies between tokens, Vision Transformer (ViT) \cite{dosovitskiy_image_2020} has been introduced to the vision community as the first pure transformer architecture that achieves comparable results with the state-of-the-art CNNs on various benchmark datasets \cite{deng_imagenet_2009,krizhevsky_learning_2012,nilsback_automated_2008}. However, ViTs lack the inductive biases present in CNNs limiting their performance on small datasets. Thus, ViT was initially trained on a large-scale JFT300M dataset \cite{sun_revisiting_2017} to be able to achieve comparable results with CNNs when finetuned on smaller datasets (e.g., ImageNet). Later, DeiT \cite{touvron_training_2021} introduces a teacher-student knowledge distillation strategy suitable for transformers through adding an additional distillation token and explores efficient augmentation and regularization techniques for training ViTs. Other approaches \cite{chen_dearkd_2022,zhao_cumulative_2023}, effectively extract inductive biases from intermediate layers of a CNN teacher and utilize them as supervisory signals for the transformer student. Further, inspired by the inherent hierarchical structure of convolutional neural networks (CNNs), researchers have explored transformer architectures that incorporate pyramid and hierarchical designs to enhance transformer performance on downstream tasks. PVTv1 \cite{wang_pyramid_2021} introduces a vision transformer with a shrinking pyramid design, while reducing the computational cost and enhancing performance on various dense predictions tasks. Swin Transformer \cite{liu_swin_2021} introduces shifted local window attention for capturing long-range dependencies with linear complexity. CrossViT \cite{chen_crossvit_2021} uses dual transformer branches for processing raw image patches with different resolutions and uses cross-attention for exchanging knowledge between both branches. \\
\newline
\textbf{Combining CNNs and Transformers:} Several recent works unravel how to effectively integrate the complementary strengths of ConvNets and Transformers into a unified hybrid model. Conformer \cite{peng_conformer_2021} employs dual structure network of CNN and Transformer for local and global features fusion. CvT \cite{wu_cvt_2021}, LeViT \cite{graham_levit_2021} and {ViT}$_{C}$ \cite{xiao_early_2021} uses early convolution to replace the regular ViT's patch embedding for better feature representation. MobileFormer \cite{chen_mobile-former_2022} introduces a lightweight network that parallelizes MobileNetV2 \cite{sandler_mobilenetv2_2018} and ViT \cite{dosovitskiy_image_2020} with a two-way bridge connection enabling bidirectional local-global feature fusion. PVTv2 \cite{wang_pvt_2022} reduces the computational cost over PVTv1 \cite{wang_pyramid_2021} by employing a linear spatial reduction attention. In addition, it uses an overlapping patch embedding and convolutional feed-forward network for enhanced feature representation. MobileViT \cite{mehta_mobilevit_2021} introduces a light-weight MobileViT block that combines convolution with transformer for modeling local-global representations. MobileViTv2 \cite{mehta_separable_2022} replaces the multi-headed self-attention (MHA) in MobileViT with separable self-attention for linear complexity. CoAtNet \cite{dai_coatnet_2021} sequentially utilizes MBConv blocks and transformer blocks with relative attention. MOAT \cite{yang_moat_2022} effectively merges the MBConv block along with the self-attention in one block rather than using them sequentially.
FasterViT \cite{hatamizadeh_fastervit_2024} utilizes traditional convolutional blocks in the initial stages and introduces a lightweight module termed Hierarchical Attention block in the later stages. \\
\newline
Building on existing hybrid approaches, our work introduces CTRL-F, which combines lightweight CNN blocks with a Multi-level Feature Cross Attention Transformer module. This design fuses local responses from the CNN and global responses from the transformer using dual fusion approaches, resulting in robust generalization even with limited data. Since CNNs include inductive biases which make it the preferred choice on limited data and Vision transformers include powerful processing capabilities, we combine benefits of both approaches. Moreover, by applying cross attention to high-level CNN features across multiple scales, CTRL-F captures richer inter-scale semantic relationships rather than relying solely on self-attention within individual feature scales.

\section{Methodology}\label{sec3}
This paper introduces an efficient lightweight hybrid design, leveraging the benefits of CNNs (i.e., inductive biases and faster convergence) and Transformers (i.e., larger model capacity, global processing, and input-adaptive weighting). In this section, we reveal how to combine the strengths of both architectures in one unified network with an effective and simple design.

\subsection{Overview}
\textbf{MBConv Block.} is one of the most widely used blocks recognized for its efficiency, lightweight design, and is sometimes termed as "inverted residual block" \cite{sandler_mobilenetv2_2018}. MBConv block initially expands the input feature channels with a factor of 4 through a 1x1 convolution, then a 3x3 depthwise convolution is applied to efficiently grasp the spatial interactions. A Squeeze-Excitation (SE) block \cite{hu_squeeze-and-excitation_2018} is then employed to extract per-channel global information for channel-wise features adaptive recalibration. Finally, a 1x1 convolution is used to project back the features into the original size by reducing the channels with a factor of 4. Batch normalization is applied after each convolution layer and GeLU activation is used. Furthermore, a residual connection is incorporated for enabling the flow of information between the input and the output. MBConv block is depicted in Figure 1.

\begin{figure}[h]
\centering
    \includegraphics[width=0.7\linewidth]{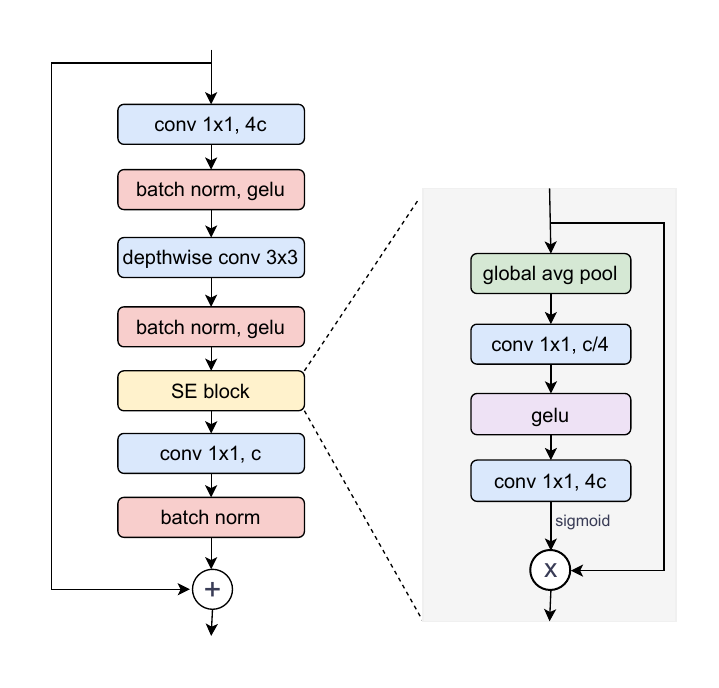}
    \caption{MBConv block with Squeeze-and-Excitation.}
\end{figure}

\noindent \textbf{Transformer Block.}  It is used as a core building block in transformer models \cite{vaswani_attention_2017}. It excels at capturing long-range dependencies and relationships between image patches. The transformer encoder consists of a stack of encoder blocks, each with two sub-layers: multiheaded self-attention (MSA) and feed-forward network (FFN). The MSA computes the pairwise similarities between image patches across multiple heads, where each head focuses on capturing different aspects of the relationships between patches. The FFN consists of two fully connected layers separated by a GeLU \cite{hendrycks_gaussian_2023} non-linearity. The first fully connected layer expands the input dimension with a factor \(r\), and the second layer shrinks the input dimension back to the original dimension. Residual connection is applied after each block, and layer normalization (LN) is applied before each block. The transformer processing blocks can be expressed as follows:

\begin{equation} 
\begin{split}
\label{eq:transformer}
    \mathbf{X}_0 & = \{  \mathbf{X}_{cls} \: \vert \vert \: \mathbf{X}_{\text{patch}_0}
     \; ... \; \mathbf{X}_{\text{patch}_n}  \} + \mathbf{X}_{pos} \\
     \mathbf{\acute{X}}_l & = MSA(LN(\mathbf{X}_{l-1})) + \mathbf{X}_{l-1} \\
     \mathbf{X}_l & = FFN(LN(\mathbf{\acute{X}}_l)) + \mathbf{\acute{X}}_l
\end{split}
\end{equation}
\(\mathbf{X}_{\text{cls}} \in \mathbb{R}^{1 \times D}\) represents the classification token (CLS) and \(\mathbf{X}_{\text{patches}} \in \mathbb{R}^{N \times D}\) represents the patch tokens, where \(\mathbf{X}_{\text{patches}} = [\mathbf{X}_{\text{patch}_0} \; ... \; \mathbf{X}_{\text{patch}_n}]\). Additionally, \(\mathbf{X}_{\text{pos}} \in \mathbb{R}^{(N+1) \times D}\) represents the positional embedding. \(N\) and \(D\) denote the number of patches and the embedding dimension, respectively.

\subsection{CTRL-F Architecture}
Our lightweight and efficient hybrid model utilizes the MBConv block as its primary convolution block. This choice is driven by the MBConv block's effectiveness and lightweight design, which aligns very well with the transformer blocks due to their shared "inverted bottleneck" design with the FFN module in transformers \cite{dai_coatnet_2021,yang_moat_2022}. Additionally, the incorporation of the Squeeze-and-Excitation (SE) module within MBConv blocks enables the convolution to capture global context by dynamically recalibrating the feature map channels according to their significance based on the information aggregated along each channel. The model architecture is structured with five stages (S0:S4) in the convolution path. In the initial stem stage (S0), a 3x3 convolution operation is performed, followed by batch normalization and a GeLU non-linearity. The subsequent stages (S1:S4) comprise a sequence of MBConv blocks, where the first block in each stage employs a strided depthwise convolution to efficiently reduce the feature map's spatial resolution by a factor of 2. Additionally, the number of channels is doubled in each stage. At the end of the convolution path, global average pooling and fully connected layers are employed to produce the final CNN prediction. \\
\indent In our design we focus how to optimally combine the merits of convolution and transformer in a unified network. To this end, we introduce a Multi-level Feature Cross-Attention Transformer (MFCA) module that processes feature maps extracted from the CNN pathway. The MFCA module resembles the dual-branch transformer structure of CrossViT \cite{chen_crossvit_2021}, but it operates on high-level feature maps generated at different convolution stages instead of operating on original image pixels. MFCA enriches the network with higher model capacity and global processing ability by utilizing two branches, each comprising a sequence of transformer encoders for capturing global representations. These branches complement each other by using a cross-attention block to integrate the obtained knowledge from each branch. The classification tokens (CLS) of the two transformer branches are summed lastly to produce the transformer prediction, which is subsequently fused with the CNN prediction using a knowledge fusion module. As a result, the proposed training paradigm enables the model to leverage the intrinsic CNN inductive biases for better generalization and faster convergence in the early training stages, while also incorporating the transformer's global processing capabilities. The overall architecture of CTRL-F is illustrated in Figure 2.
\begin{figure}[tb!]
\centering
    \includegraphics[width=\linewidth]{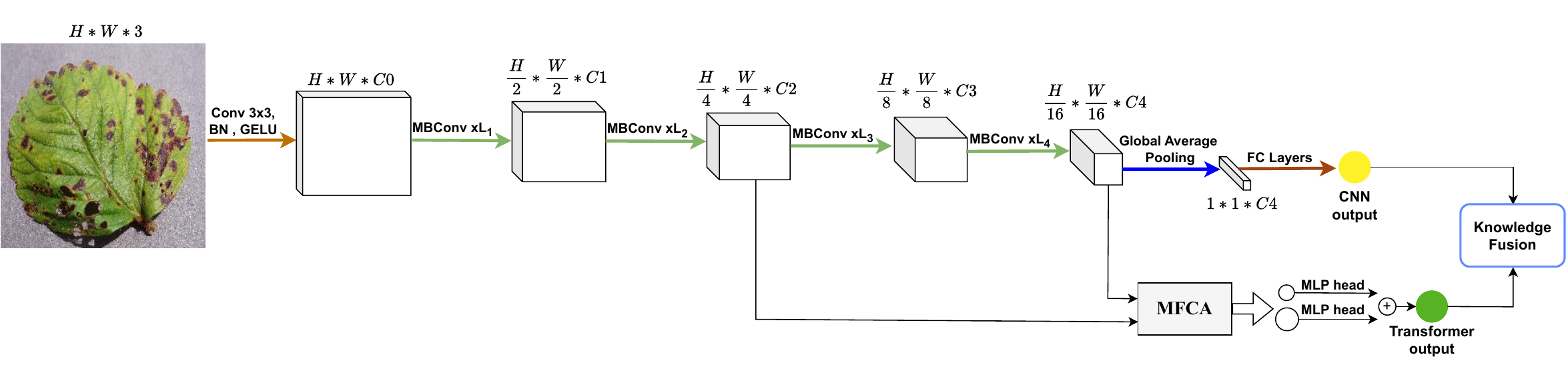}
    \caption{\textbf{Overview of our proposed CTRL-F.} The input image is fed to the convolutional path, which comprises a sequence of hierarchical stages including a stem block in the first stage (S0) and MBConv blocks in the remaining stages (S1:S4). MFCA module processes two features' representations extracted from an intermediate convolutional stage (S2) and the last convolutional stage (S4), respectively. The local responses acquired from the convolutional path are effectively combined with the global responses acquired from the MFCA module via knowledge fusion strategies for improved predictive capabilities.}
\end{figure}

\subsection{Multi-level Feature Cross-Attention Transformer (MFCA) module}
The MFCA module depicted in Figure 3, utilizes a dual-branch transformer structure like CrossViT \cite{chen_crossvit_2021} but the main difference that it operates on high-level feature maps produced from the CNN pathway. The MFCA module take advantage of operating on high level, low-resolution feature maps, serving two main purposes. First, it operates on enhanced feature representations, and benefits from leveraging the inductive biases acquired through convolution. Second, it minimizes the computational overhead by processing low-resolution feature maps with better semantic information, avoiding the high complexity of directly processing the original image pixels, making the network feasible in practice for real-time use.
\newline
\indent The impact of patch size choice on transformer in terms of efficiency and complexity is substantial. Employing small and fine-grained image patches as input yields to significant performance gains \cite{wang_pyramid_2021,chen_crossvit_2021}, but comes with higher computations and latency, while using coarse-grained patches as input leads to much fewer FLOPs and latency, but negatively affects the model performance \cite{dosovitskiy_image_2020}. Within our MFCA module, each transformer branch independently operates on patches of varying sizes extracted from features obtained at different convolution stages. This design enables both branches to effectively process fine-grained patches with much reduced complexity by operating on feature-level representations instead of the pixel-level representation. Specifically, the small and large branches operate on patches of sizes 2 and 8, respectively. Our approach leverages the benefits of using fine-grained patches in both branches with minimal impact on the computational requirements, as the feature maps resolution obtained at stage \(j\) (\(\Psi_j\)) is smaller than the feature maps resolution obtained at stage \(i\) (\(\Psi_i\)), where stage \(j\) follows stage \(i\), and both resolutions are significantly smaller than the original image resolution (\(\Psi_j \ll \Psi_i \ll I\)), where \(\Psi_j \in \mathbb{R}^{\frac{H}{2^j} \times \frac{W}{2^j} \times C_j}\), \(\Psi_i \in \mathbb{R}^{\frac{H}{2^i} \times \frac{W}{2^i} \times C_i}\), and \(I \in \mathbb{R}^{H \times W \times 3}\).

\indent The large branch within the MFCA module processes larger patches with N transformer encoders and larger embedding dimensions, whereas the small branch processes smaller patches with M transformer encoders and smaller embedding dimensions. The global feature representations obtained from each branch are then exchanged and fused L times using cross-attention block to exploit the global knowledge gained from both branches. Finally, the CLS tokens obtained from both branches are projected to a vector of size equal to the number of classes and added together to present the final prediction of the MFCA module.
\begin{figure}[tb!]
\centering
    \includegraphics[width=\linewidth]{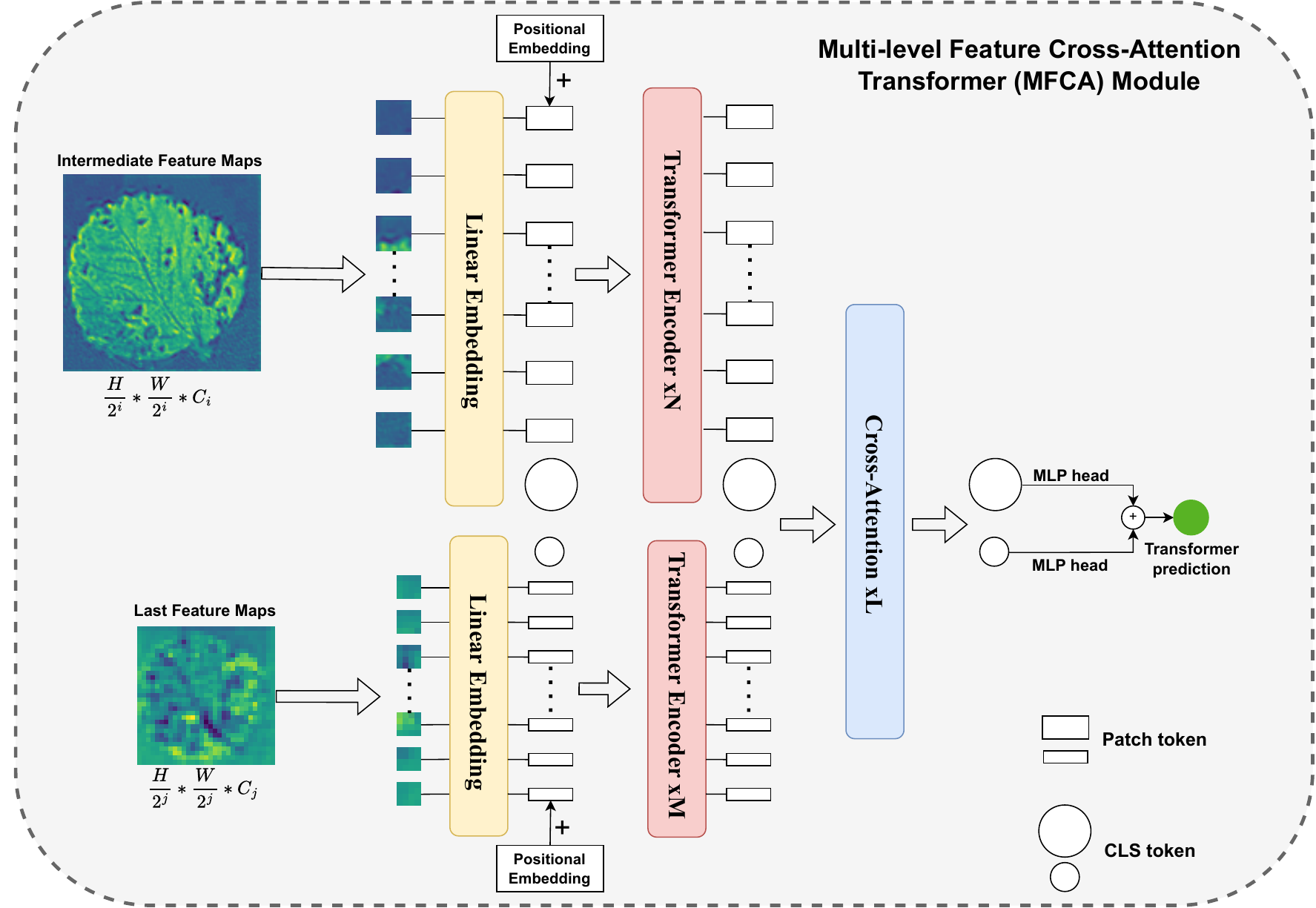}
    \caption{\textbf{Overview of our MFCA module.} This module processes two high-level features obtained at different convolutional stages using a sequence of transformer encoders. The global knowledge obtained from both branches are then exchanged multiple times using consecutive cross-attention blocks.}
\end{figure}

\subsubsection{Cross-Attention Block}
Our cross-attention block for the large branch is depicted in Figure 4. The CLS token in each branch within the MFCA module acts as a global summary for image features derived from various image patches interactions. The cross-attention block exchanges the knowledge of one branch represented in the CLS token with the patches of the other branch, enriching each branch with knowledge aggregated across different scales and representations. For the large branch, the CLS token is projected to the same embedding size of the small branch tokens via a linear layer, ensuring compatible alignment with the smaller branch's token embeddings. The aligned CLS token from the large branch are then concatenated with the small branch tokens. This combined set of tokens is then subjected to multi-headed cross-attention, allowing the exchange of knowledge between both branches. Finally, the large branch's CLS token is projected back to its original dimension and concatenated with the remaining large branch tokens. The same procedure is mirrored for the small branch, where its CLS token is aligned and interacts with the large branch's tokens in the multi-headed cross-attention step. As depicted in Figure 4, given the input tokens for the large branch \( X^l = \{ \mathbf{X}_{\text{cls}}^l \, || \, \mathbf{X}_{\text{patch}_0}^l \ldots \mathbf{X}_{\text{patch}_n}^l \} \) and \( X^s = \{ \mathbf{X}_{\text{cls}}^s \, || \, \mathbf{X}_{\text{patch}_0}^s \ldots \mathbf{X}_{\text{patch}_n}^s \} \) for the small branch. We obtain the new embedding tokens for the large branch as follows:

\begin{equation} 
\begin{split}
\begin{aligned}
\label{eq:cross_attention}
     & \mathbf{X}_{cls}^{\acute{l}}  = LN(Linear(\mathbf{X}_{cls}^{l})) \\
     & \mathbf{X}^{fused}  = {\{\mathbf{X}_{cls}^{\acute{l}} \vert \vert \mathbf{X}_{\text{patch}_0}^{s} \; ... \; \mathbf{X}_{\text{patch}_n}^{s} \}} \\ 
     & \mathbf{X}_{cls}^{l+1}  = Linear(MCA({X}^{fused}) + \mathbf{X}_{cls}^{\acute{l}})) \\
     & \mathbf{X}^{l+1}  = {\{\mathbf{X}_{cls}^{l+1} \vert \vert \mathbf{X}_{\text{patch}_0}^{l}
     \; ... \; \mathbf{X}_{\text{patch}_n}^{l}\}}
\end{aligned}
\end{split}
\end{equation}
The \(\mathbf{X}_{cls}^{l}\) is projected for dimension alignment to \(\mathbf{X}_{cls}^{\acute{l}}\) through a linear layer. Consequently, layer normalization (LN) is applied to the aligned CLS token. Where the linear layer in the large branch maps the original CLS token to the dimensions of the other branch tokens \(dim(\mathbf{X}_{cls}^{\acute{l}}) = dim(\mathbf{X}^{s})\). The obtained CLS token is then concatenated with the small branch tokens \((\mathbf{X}^{fused}\)) and subjected to multi-headed cross attention (MCA). A residual connection is then applied and the new CLS token \((\mathbf{X}_{cls}^{l+1})\) is obtained through projecting it back to the dimension of the large branch tokens through a linear layer. Finally, the new CLS token is concatenated with the original tokens of the large branch to obtain the new tokens for the upcoming cross-attention block \((\mathbf{X}^{l+1})\). The MCA can be expressed as
\begin{equation} 
\begin{split}
\begin{aligned}
\label{eq:MCA}
     & \mathbf{Q} = \mathbf{W}_{q} \, \mathbf{X}_{cls}^{\acute{l}}, \ \ 
     \mathbf{K} = \mathbf{W}_{k} \, \mathbf{X}^{fused},  \ \
     \mathbf{V} = \mathbf{W}_{v} \, \mathbf{X}^{fused} \\
     & MCA(\mathbf{Q},\mathbf{K},\mathbf{V}) = Softmax(\frac{\mathbf{Q\mathbf{K}^{\text{T}}}}{\sqrt{\mathbf{d}_{k}}}) \mathbf{V}
\end{aligned}
\end{split}
\end{equation}
where \( W_q, W_k, W_v \in \mathbb{R}^{c \times d_k} \) are the learnable weight matrices, \( c \) and \( d_k \) are the tokens embedding dimension and the dimension per head.

\begin{figure}[tb!]
\centering
    \includegraphics[width=\linewidth]{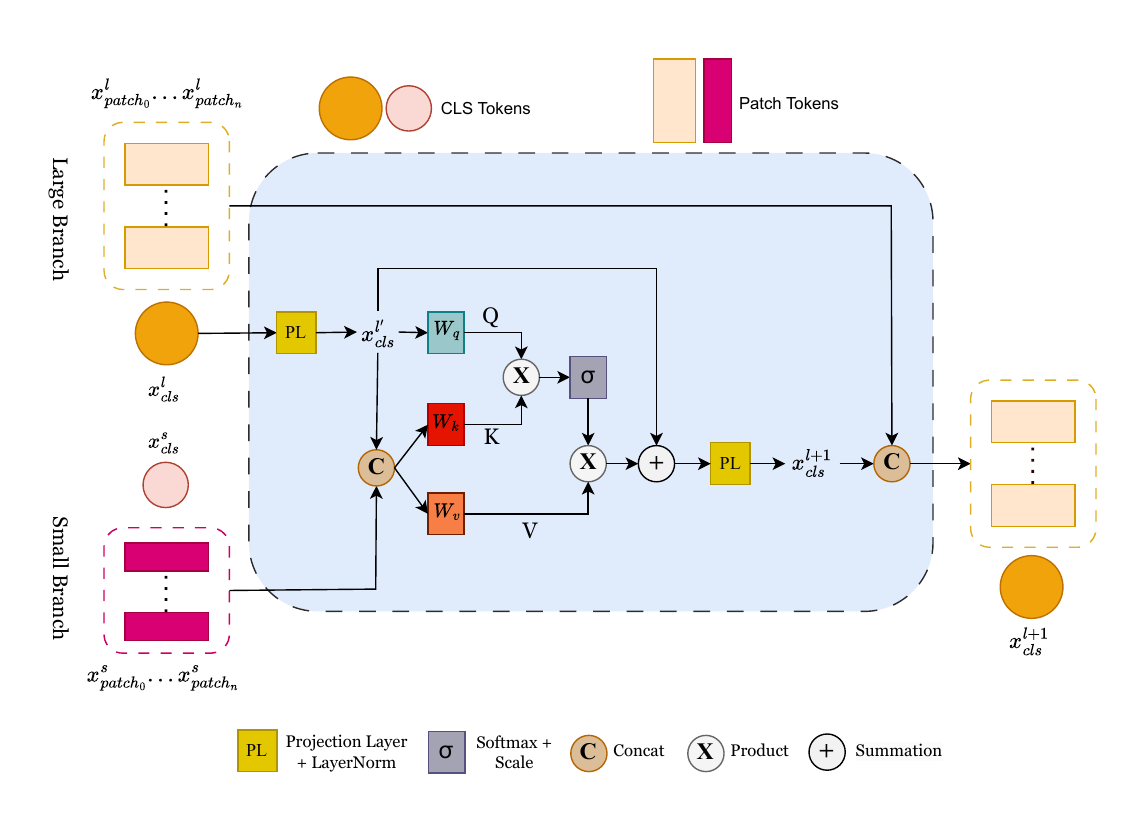}
    \caption{\textbf{Cross-Attention block for the large branch.} The classification token of the large branch \(\mathbf{X}_{cls}^{l}\) is aligned to the embedding dimension of the small branch tokens through a linear layer. It acts as a query token for the cross-attention block to interact with the key and value generated from the small branch embedding tokens through multi-headed cross-attention (MCA). The resulting cross-representation embedding features from MCA are then projected back to the dimensions of the large branch tokens and concatenated with the tokens of the large branch. The procedure for the small branch mirrors that of the large branch, but the query token of the small branch interacts with the tokens of the large branch in the cross-attention block.}
\end{figure}

\subsection{Knowledge fusion}
To effectively integrate the local and global information learned from both branches (i.e., convolutional path and MFCA module), we introduce two knowledge fusion techniques named adaptive knowledge fusion and collaborative knowledge fusion. These proposed techniques are designed to guarantee that the knowledge gained from both branches significantly contributes to the final prediction of the model, avoiding any noticeable bias towards a particular branch during the training process.
\subsubsection{Adaptive knowledge fusion}
 The proposed adaptive knowledge fusion technique is depicted in Figure 5. When using this fusion technique, we let the final linear layer in the CNN branch and MFCA module produce a vector with dimensions corresponding to the number of classes (\( c \)). The two output vectors representing the CNN prediction (\( \tilde{y}_{\text{cnn}} \in \mathbb{R}^{1 \times c} \)) and the transformer prediction (\( \tilde{y}_{\text{trans}} \in \mathbb{R}^{1 \times c} \)) undergo L1-normalization. Subsequently, the normalized CNN prediction is fused with the normalized transformer prediction using an adaptive weighting hyperparameter (\( \lambda \)) to obtain the final prediction of our proposed network. The lambda hyper-parameter is dynamically adapted to prioritize the CNN prediction in the initial epochs and as the training progress, it begins increasingly to pay more attention for the global responses by adapting the lambda gradually to focus more on the transformer prediction. Prioritizing the CNN prediction in the initial training stages also ensures that the MFCA module receives high-quality feature maps, enabling the transformer to effectively extract and utilize global context. It's noteworthy to mention that it was observed that without applying L1-normalization to both output vectors, the model tends to favor the CNN prediction by boosting the magnitude of its vector elements heavily over the transformer prediction leading to diminishing the impact of transformer during training. This challenge arises because the model, in its pursuit of immediate optimization gains, leans heavily towards following the guidance from the CNN and doesn't account for the long-term contribution of the transformer which will lead to  better global optimization at the late training stages. Hence, normalizing both branches' predictions is crucial as it forces the model to learn from both branches since both vectors are mapped to comparable range of values (-1 to 1), ensuring that the final prediction can benefit from both branches' capabilities. The final prediction is given as follows:

\begin{equation} 
\begin{aligned}
\label{eq:AKF}
     & \tilde{y} = Softmax((\lambda \times \vert \vert \tilde{y}_{\text{cnn}} \vert \vert_1 + (1- \lambda) \times \vert \vert \tilde{y}_{\text{trans}} \vert \vert_1) \times \alpha)
\end{aligned}
\end{equation}
where $\lambda$ is an adaptive weighting hyperparameter that decreases uniformly over the number of training epochs (i.e., 0.7 → 0.3), and $\alpha$ is a scaling hyperparameter that enlarges the fused normalized vector for better convergence.

\begin{figure}[h]
\centering
    \includegraphics[width=0.95\linewidth]{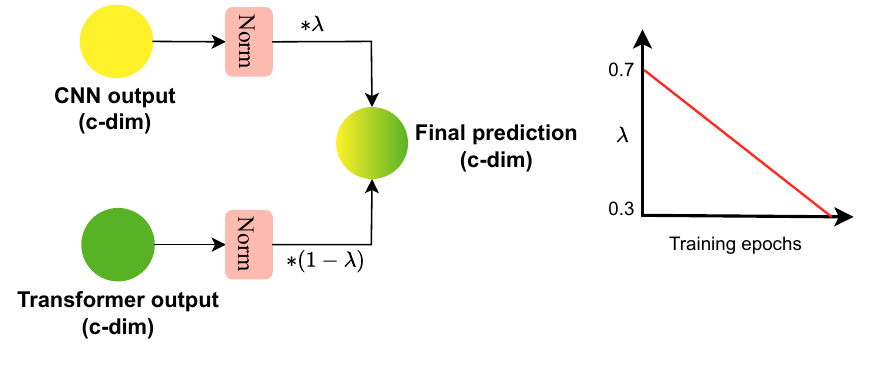}
    \caption{Adaptive knowledge fusion strategy}
\end{figure}

\subsubsection{Collaborative knowledge fusion}
The proposed collaborative knowledge fusion technique is depicted in Figure 6. Given the output vector of the CNN branch (\( \tilde{y}_{\text{cnn}} \in \mathbb{R}^{1 \times n} \)) and the MFCA module branch (\( \tilde{y}_{\text{trans}} \in \mathbb{R}^{1 \times m} \)), each vector is passed through a linear layer to align them to a common dimension \( k \). The resulting vectors are then concatenated into a single vector (\( \tilde{y}_{\text{fused}} \in \mathbb{R}^{1 \times 2k} \)), to effectively capture information from both branches. Subsequently, a dropout layer \cite{srivastava_dropout_2014} is then applied, followed by a linear layer which transforms the concatenated vector to a dimension with number of elements corresponding to the number of classes (\( \tilde{y} \in \mathbb{R}^{1 \times c} \)). The dropout layer role is crucial in this fusion module as it ensures that the final prediction doesn't rely on specific neurons belonging to a certain branch output. Without adding dropout to the fusion module, the model can easily favor one branch prediction by boosting its neurons outputs while diminishing the influence of the other branch, hence adding dropout forces the model each time to minimize the error with different random set of neurons coming from both branches, as well as reduce overfitting. The final prediction is given as follows:

\begin{equation} 
\begin{aligned}
\label{eq:CKF}
     & \tilde{y} = Softmax(Linear_c (Dropout(Concat(Linear_k(\tilde{y}_{\text{cnn}}), Linear_k(\tilde{y}_{\text{trans}})))))
\end{aligned}
\end{equation}

Where $Linear_k$, $Linear_c$ represents the linear layers that aligns the given feature vector to a corresponding vector with \(k\) and \(c\) dimension, respectively. In our experiments we fix the dropout ratio $(\alpha)$ to 0.5.

\begin{figure}[h]
\centering
    \includegraphics[width=0.95\linewidth]{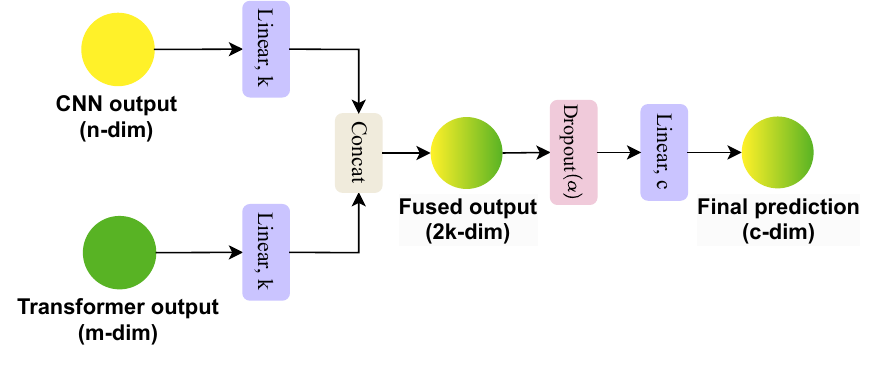}
    \caption{Collaborative knowledge fusion strategy}
\end{figure}

\section{Experiments}
In this section, we perform extensive experiments to demonstrate the efficiency of our proposed novel CTRL-F models compared to some state-of-the-art Convolutional Neural Nets (ConvNets), Vision Transformers (ViTs), and hybrid models discussed in the related work section across multiple datasets.
\subsection{Experimental Setup}
\textbf{Datasets.} Datasets. We conducted our experiments on three benchmarks datasets: PlantVillage \cite{mohanty_using_2016}, Oxford-102 Flowers \cite{nilsback_automated_2008} and CUB-200 \cite{WahCUB_200_2011} datasets. The PlantVillage dataset is widely recognized as the largest open-source dataset for classifying various plant leaf diseases. It consists of 54303 healthy and diseased leaf images belonging to 38 categories from 14 distinct plant crop species. For training, 80\% of images from each category were used, while the remaining 20\% are used for evaluation. The Oxford-102 Flowers dataset comprises 102 unique flower categories, each containing between 40 and 258 images. The training set consists of only 20 images per class, while the remaining images are used for evaluation. The CUB-200 dataset is a widely used benchmark for fine-grained visual categorization. It consists of 200 unique bird species, with each class containing approximately 60 images, split into $\sim30$ for training and $\sim30$ for testing. Both Oxford-102 and CUB-200 datasets pose a significant challenge due to the limited number of images available for training in each class. The distribution of training and testing sets for each dataset is given in Table 1. 

\begin{table}[h]
\centering
\caption{Datasets specifications.}
\begin{tabular}{ l | ccc}
\toprule
 Dataset  & Train Size & Test Size & \#classes \\
 \midrule
 PlantVillage \cite{mohanty_using_2016} & 43,444 & 10,861 & 38 \\ 
 Oxford-102 \cite{nilsback_automated_2008}   & 2,040  & 6,149  & 102 \\
 CUB-200 \cite{WahCUB_200_2011}      & 5,994  & 5,794  & 200 \\
\bottomrule
\end{tabular}
\end{table}

\noindent \textbf{Implementation Details.} The proposed CTRL-F models are trained from scratch on PlantVillage,  Oxford-102 and CUB-200 datasets. Our models are trained for 100 epochs on PlantVillage dataset and 200 epochs on Oxford-102 and CUB-200 datasets. We utilize an image resolution of 224 \(\times\) 224 and employ a batch size of 32 during the training process. The models are trained using PyTorch on NVIDIA P100 GPU using AdamW optimizer \cite{loshchilov_decoupled_2018} with a cosine decay learning rate scheduler. We used an initial learning rate of 5e-4, a minimum learning rate of 1e-5, a weight decay of 0.05 and 5 epochs of linear warm-up. In addition, basic data augmentation (e.g., horizontal flipping and rotation) are used during training. It is noteworthy to mention that the learning rate parameters should be carefully selected to ensure balanced training of both branches (i.e., CNN head and transformer head), preventing one from dominating while the other vanishes. \\
\newline
\noindent \textbf{Variants.} We build two variants of our proposed CTRL-F network, named CTRLF-B and CTRLF-S. The CTRLF-B is considered the base variant, while the CTRLF-S is considered the smaller and lighter variant, with less than 0.5\(\times\) of the base variant size and learnable parameters. The architectural hyper-parameters of the model variants are given in Table 2. We use N = 3 and M = 3 for both variants, where N and M denotes the number of transformers blocks in the large and small branch of the MFCA module respectively, and L denotes the number of cross attention blocks. The number of heads is the same in the transformer blocks of both branches and cross-attention blocks (h = 6). The expansion ratio in the feed-forward network of the transformer block is fixed in both variants (r = 12 and 4 for the small branch and large branch, respectively). 

\renewcommand{\arraystretch}{1.1}
\begin{table}[h]
\centering
\caption{Variants of CTRL-F model.}
\begin{tabular}{ p{1.8cm} | c c |ccccc}
\hline
 \multirow{3}{*}{Model} & \multicolumn{2}{c|}{Convolution branch}  & \multicolumn{5}{c}{MFCA module} \\ 
 \cline{2-8}
 & \multirow{2}{*}{\makecell{ \#MBConv \\ per stage}} & \multirow{2}{*}{ \makecell{MBConv \\ depth per stage}} & \multicolumn{2}{c}{Patch size} & \multicolumn{2}{c}{Embedding dim} & L \\ 
 & & & Small & Large & Small & Large &  \\
 \hline
 CTRLF-S & \{2,2,3,5\} & \{32,64,128,256\} & 2 & 8 & 128 & 256 & 2\\
 CTRLF-B & \{2,2,4,8\} & \{64,92,196,256\} & 2 & 8 & 192 & 384 &	4\\
 \hline
\end{tabular}
\end{table}


\subsection{Ablation Studies}
\textbf{Comparison of the proposed knowledge fusion strategies.} We compare our small (CTRLF-S) and base (CTRLF-B) variants under the proposed adaptive knowledge fusion (AKF) and collaborative knowledge fusion (CKF) strategies (see section 3.4). Both fusion strategies demonstrate superior performance with the proposed variants when trained from scratch on both datasets. Table 3 presents the top-1 accuracy and the number of parameters for each variant when applying each strategy. For the Oxford-102 dataset, employing the adaptive knowledge fusion strategy with the base variant (CTRLF-B + AKF) achieves top-1 accuracy of 84.01\%, surpassing its collaborative knowledge fusion counterpart (CTRLF-B + CKF) with an absolute gain of 1.01\%. Whereas, employing the adaptive knowledge fusion strategy with the small variant (CTRLF-S + AKF) achieves top-1 accuracy of 82.42\%, surpassing its collaborative knowledge fusion counterpart (CTRLF-S + CKF) with an absolute gain of 0.54\%. For the CUB-200 dataset, employing the CTRLF-B + CKF outperforms CTRLF-B + AKF with an absolute gain of 0.91\%, while the CTRLF-S + CKF outperforms CTRLF-S + AKF with an absolute gain of 1.08\%. However, for the PlantVillage dataset, the CTRLF-B + CKF outperforms CTRLF-B + AKF with an absolute gain of 0.02\%, while the CTRLF-S + CKF outperforms CTRLF-S + AKF with an absolute gain of 0.06\%. Despite the base variant has about \( 2 \times \) number of parameters and FLOPs compared to the small variant, it is observed that the base variants outperform the small variant when using AKF and CKF knowledge fusion techniques on the three datasets, which proves the scalability of our CTRL-F.

\renewcommand{\arraystretch}{0.95}
\begin{table}[h]
    \centering
    \caption{Comparison of the proposed knowledge fusion strategies when employed with our variants.}
    \begin{tabular}{ cccccc}
        \toprule
        Variant   & Knowledge & Params & Oxford-102 & CUB-200 & PlantVillage \\ 
        & fusion    &        & top-1 acc  & top-1 acc & top-1 acc \\
                               
        \midrule
        CTRLF-S & AKF & 9.97M & 82.42\% & 56.27\% & 99.85\% \\
        CTRLF-S & CKF & 9.99M & 81.88\% & 57.35\% & 99.91\% \\
        CTRLF-B & AKF & 21.36M & 84.01\% & 60.77\% & 99.89\% \\
        CTRLF-B & CKF & 21.39M & 83.00\% & 61.68\% & 99.91\% \\
        
        \bottomrule
    
    \end{tabular}
\end{table}

\noindent \textbf{Investigating the individual contribution of each branch after training the entire architecture.} Since the adaptive knowledge fusion (AKF) strategy combines the convolutional path prediction with the MFCA module prediction using an adaptive weighting hyperparameter (\(\lambda)\) to obtain the networks' final prediction, we can easily investigate the impact of each branch prediction on its own and how their ensemble enhances the whole network predictive capabilities. Table 4 reports the top-1 accuracy for AKF strategy on the experimented datasets for the small variant when using the CNN head prediction only and when using the MFCA module head prediction only, and when combining both using the proposed adaptive knowledge fusion strategy. It can be observed that combining both predictions yields favorable performance, ensuring that the model can benefit from effectively fusing local-global representations.

\begin{table}[h]
    \centering
    \caption{Comparing the individual contribution of each branch against the fused weighted combination of both branches for the small variant of CTRL-F.}
    \begin{tabular}{ l | cccc}
        \toprule
        Dataset   & CNN    & MFCA module & AKF \\ 
                  & branch & branch      &     \\
                               
        \midrule
        Oxford-102   & 80.21\% & 75.14\% & 82.42\%  \\
        CUB-200      & 43.32\% & 26.56\% & 56.27\%  \\
        PlantVillage & 99.70\% & 98.09\% & 99.85\%  \\
        \bottomrule

    \end{tabular}
\end{table}

\vspace{0.5em}

\noindent \textbf{Patch Size analysis.} Since we utilize two independent transformer encoders to process two convolutional feature maps (i.e., intermediate, and final features). We conduct experiments to investigate the impact of changing the patch sizes on both performance and complexity. Specifically, we tested the small variant of our CTRL-F with different pairs of patch sizes that are (8,2), (2,2), (8,7), (14,2). We observe that employing smaller patch sizes (i.e., 2) for the final feature maps yield in better results as it operates on finer and more-detailed patches without significantly increasing FLOPs count. This is particularly because the resolution of feature maps obtained at the final stage is substantially smaller than the feature maps obtained at an intermediate stage (see Figure 2). Conversely, our findings from Table 5 indicate that employing a very small patch size (i.e., 2) for the feature maps obtained at an intermediate stage (i.e., stage 3 in our case) leads to nearly 3$\times$ higher FLOPs count compared to other experimented patch sizes as 8 and 14, despite a slight reduction in model parameters. \\
\indent In addition, the selected intermediate feature maps processed by the large transformer branch in the MFCA module has 4$\times$ resolution of the final feature maps processed by the small transformer branch. Thus, selecting a patch size of 8 for the large branch and 2 for the small branch enables each transformer encoder to process patches holding the same content but with different level of granularity. This setup is necessary for exchanging the knowledge in cross-attention between patches pointing to the same receptive field but with different representations, ensuring efficient exchange between multi-level feature representations. For instance, allowing the large branch to process patches of size 8$\times$8, and the small branch to process patches of size 2$\times$2 surpasses other choices of patch sizes in terms of accuracy when using the AKF and CKF knowledge fusion techniques.  \\
\indent We employ an input image resolution of 224$\times$224 for our CTRL-F models. Consequently, the intermediate feature maps obtained at stage 3 have a resolution of 56$\times$56, while the feature maps obtained at the last stage (i.e., stage 5) have a resolution of 14$\times$14. We tested all possible combinations of patch sizes (i.e., those divisible by the feature resolution) for the feature maps obtained at stages 3 and 5, to assess their impact on FLOPs and parameters, as shown in Figure 7 and Figure 8, respectively. We can see that processing larger patch sizes, especially for the intermediate feature maps (stage 3) significantly reduces FLOPs but greatly increases the number of parameters. However, processing larger patch sizes for the final feature maps (stage 5) has a minimal impact on parameters and FLOPs, due to the lower-resolution of these feature maps compared to the 4$\times$ higher-resolution feature maps at stage 3. Thus, choosing a moderate patch size for high-resolution feature maps (i.e., at stage 3) strikes a balance, offering reasonable computational efficiency, memory footprint and model size as depicted in Figure 7 and Figure 8.

\begin{table}[h]
    \centering
    \caption{Effect of patch sizes on performance and complexity. The red color indicates changes from the CTRLF-S variant.}
    \begin{tabular}{ cccccc}
        \toprule
        Knowledge   & \multicolumn{2}{c}{Patch size} & FLOPs & Params & Accuracy \\ 
         Fusion     & Large            & Small       &       &        &          \\
        \toprule
        \multirow{4}{*}{CKF} & 8  & 2 & 1.43G & 9.99M  & 81.88\% \\
                             & \textcolor{red}{2}  & 2 & 3.75G & 9.20M  & 80.61\% \\
                             & 8  & \textcolor{red}{7} & 1.37G & 11.46M & 80.48\% \\
                             & \textcolor{red}{14} & 2 & 1.33G & 12.15M & 81.47\% \\
                             
        \cmidrule(lr){1-6}

        \multirow{4}{*}{AKF} & 8  & 2 & 1.43G & 9.97M  & 82.42\% \\
                             & \textcolor{red}{2}  & 2 & 3.75G & 9.18M  & 82.35\% \\
                             & 8  & \textcolor{red}{7} & 1.37G & 11.44M & 81.80\% \\
                             & \textcolor{red}{14} & 2 & 1.33G & 12.12M & 82.42\% \\

        \bottomrule

    \end{tabular}
\end{table}

\begin{figure}[h]
\centering
    \includegraphics[width=\linewidth]{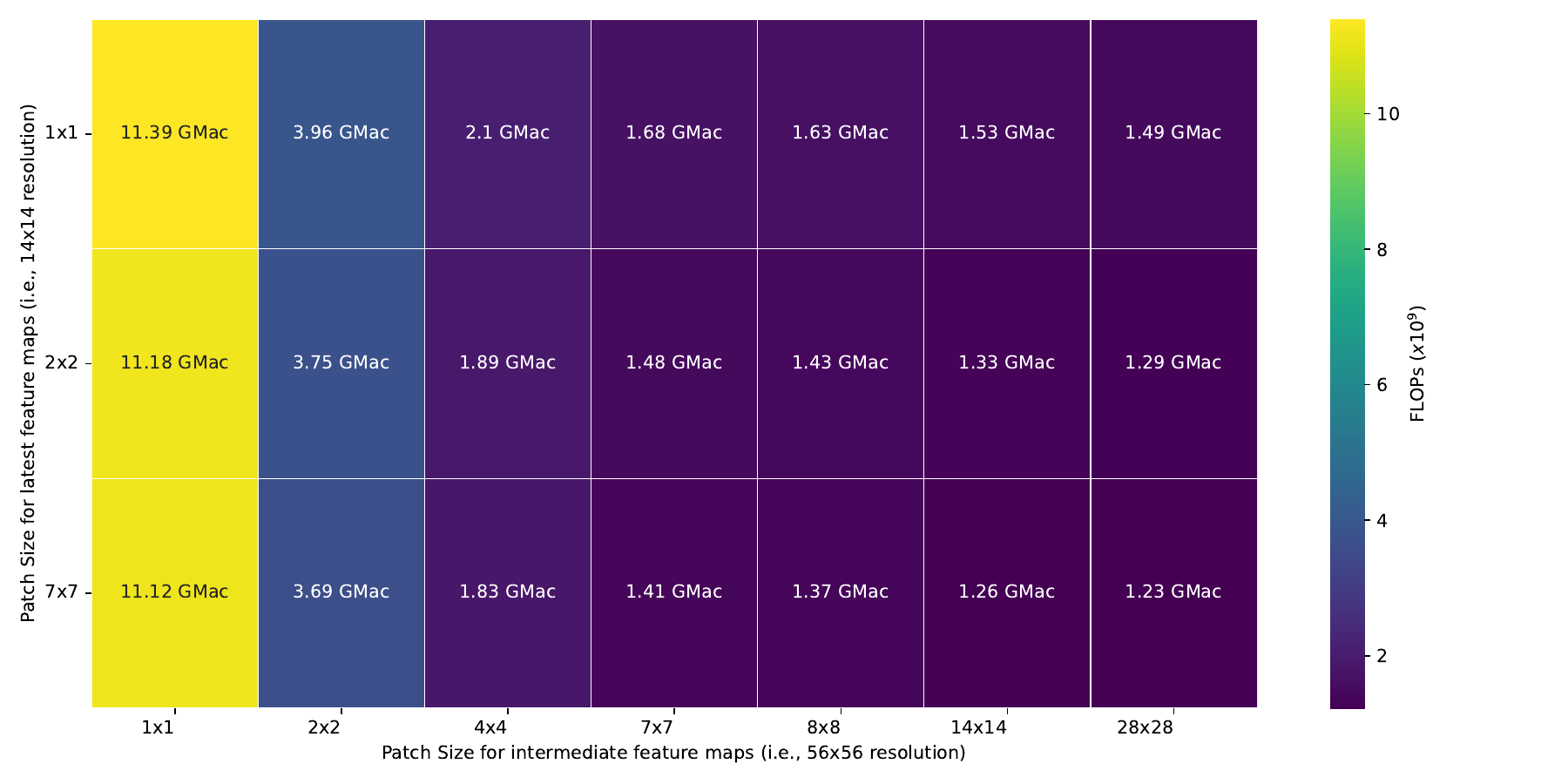}
    \caption{\textbf{Effect of changing patch sizes on FLOPs.} The heatmap illustrates the FLOPs for various valid combinations of patch sizes applied to features obtained at stages 3 and 5 of the convolution path. It is concluded that increasing the patch size strongly correlates with a decrease in FLOPs, especially when the feature resolution is relatively large (e.g., 56\(\times 56)\), demonstrating enhanced computational efficiency with larger patch sizes.}
\end{figure}

\begin{figure}[H]
\centering
    \includegraphics[width=\linewidth]{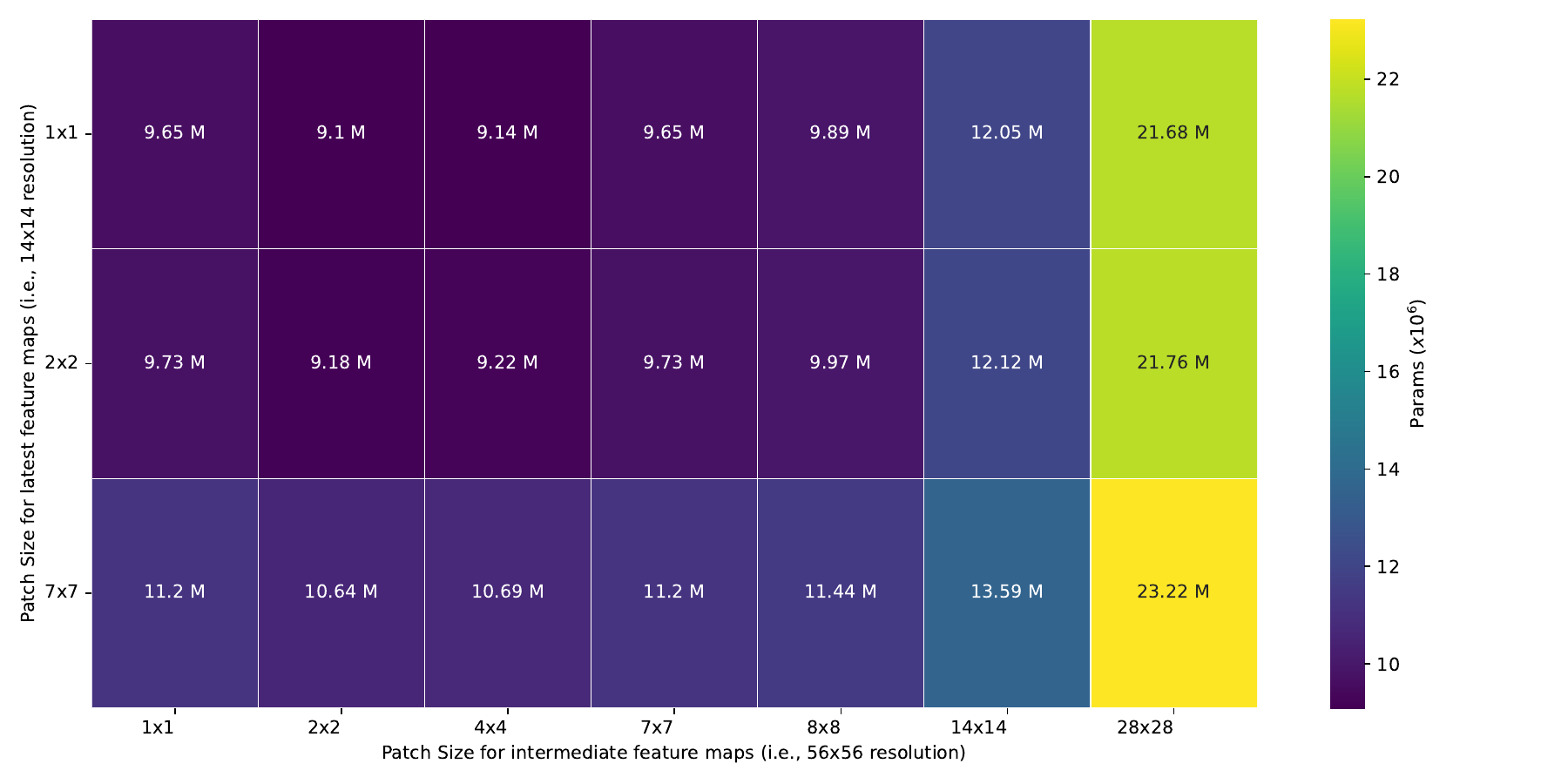}
    \caption{\textbf{Effect of changing patch sizes on parameters.} The heatmap illustrates the parameters for various valid combinations of patch sizes applied to features obtained at stages 3 and 5 of the convolution path. It is concluded that the patch size is strongly positively correlated with the parameter count, revealing that increasing patch size requires more memory footprint and increases the model size.}
\end{figure}

\noindent \textbf{Impact of image resolution on model performance.} Since Oxford-102 and CUB-200 datasets have variable high-resolution images. We investigate the effect of training the CTRL-F variants when using larger image resolution for both dataset. It is observed from Table 6 that our model demonstrates improved performance by operating at higher resolutions. This enhancement in performance underscores the robustness and adaptability of our model.

\begin{table}[h]
    \centering
    \caption{Ablation study of increasing the images resolution on Oxford-102 and CUB-200 datasets.}
    \begin{tabular}{ ccccc}
        \toprule
        Variant   & Image size & FLOPs & Oxford-102  & CUB-200 
        \\ 
               &   &   & top-1 acc & top-1 acc \\
        \toprule 
        
        \multirow{2}{*}{CTRLF-S (+CKF)} & 224$^2$  & 1.43G & 81.88\% & 57.35\% \\
                                        & 384$^2$  & 4.19G & 82.57\% & 63.19\% \\
                                        
        \cmidrule(lr){1-5}

        \multirow{2}{*}{CTRLF-S (+AKF)} & 224$^2$  & 1.43G & 82.42\% & 56.27\% \\
                                        & 384$^2$  & 4.19G & 83.77\% & 61.68\% \\
        
        \cmidrule(lr){1-5}

        \multirow{2}{*}{CTRLF-B (+CKF)} & 224$^2$  & 3.29G & 83.00\% & 61.68\% \\
                                        & 384$^2$  & 9.65G & 85.56\% & 65.48\% \\

        \cmidrule(lr){1-5}

        \multirow{2}{*}{CTRLF-B (+AKF)} & 224$^2$  & 3.29G & 84.01\% & 60.77\% \\
                                        & 384$^2$  & 9.65G & 85.88\% & 65.43\% \\
                                        
        \bottomrule

    \end{tabular}
\end{table}

\subsection{Main Results}
We seek to demonstrate the effectiveness of CTRL-F variants, evaluating their performance on both a small and challenging datasets (i.e., Oxford-102 Flowers and CUB-200 datasets), and on a diverse and large dataset (i.e., PlantVillage dataset). We compare the performance of our CTRL-F variants against some of the current state-of-the-art methods with comparable computational cost (i.e., parameters and FLOPs) belonging to the categories of pure ConvNets, Vision Transformers (ViTs), and Hybrid models. To ensure a fair comparison, all models are trained from scratch (i.e., without any pre-training) on all datasets using an input size of 224\(\times\)224. A state-of-the-art comparison in terms of top-1 accuracy, number of parameters, and FLOPs is given in Table 7. \\
\newline
\noindent \textbf{Comparison with CNNs.} We compare our proposed variants against some state-of-the-art CNN-based models with comparable Params and FLOPs. Our CTRLF-S (+AKF) surpasses lightweight CNNs including MobileNetV3-L and EfficientNet-B3 with an absolute margin of 17.0\% and 12.6\% respectively, on the Oxford-102 dataset. Our larger variant, CTRLF-B (+AKF) outperforms ConvNeXt-T and EfficientNetV2-S by 30.6\% and 20.6\% respectively, on Oxford-102 dataset. Similarly, CTRLF-S (+CKF) surpasses MobileNetV3-L and EfficientNet-B3 by 17.9\% and 9.3\%, while CTRLF-B (+CKF) surpasses ConvNeXt-T and EfficientNetV2-S by 29.9\% and 27.6\% on CUB-200 dataset. On the PlantVillage dataset, CTRLF-B (+CKF) achieves 0.32\% and 0.07\% higher accuracy than ConvNeXt-T and EfficientNetV2-S, respectively. It can be deduced that our proposed variants significantly outperform the widely used CNNs in terms of accuracy with comparable cost. Also, this reveals the efficiency of our proposed variants when trained from scratch and without any extra data on both small datasets (Oxford-102 and CUB-200) and on a larger dataset (PlantVillage). \\
\newline
\noindent \textbf{Comparison with transformers.} Due to their reliance on extensive datasets, pure transformer models often require large amounts of data to achieve state-of-the-art results. Consequently, when trained from scratch on small datasets, the performance of these transformers typically falls behind pure CNNs and hybrid models. Our variants significantly outperform recent ViTs. The CTRLF-S (+AKF) variant surpasses CrossViT-15, PVTv1-S and Swin-T with an absolute margin of 33.8\%, 29.6\% and 21.6\% on the Oxford-102 dataset respectively, and our CTRLF-S (+CKF) surpasses CrossViT-15, PVTv1-S and Swin-T by margins of 0.54\%, 0.32\% and 0.20\% on PlantVillage dataset and by 36.6\%, 39.4\% and 28.8\% on CUB-200 dataset while being smaller and more resource efficient. Additionally, our CTRLF-B (+AKF) variant surpasses CrossViT-15, PVTv1-S and Swin-T by 35.4\%, 31.2\% and 23.2\% on the Oxford-102 respectively, and CTRLF-B (+CKF) outperforms CrossViT-15, PVTv1-S and Swin-T by 0.54\%, 0.32\% and 0.20\% on the PlantVillage and by 40.9\%, 43.7\% and 33.1\% on CUB-200. \\
\newline
\noindent \textbf{Comparison with hybrid models.} Recent advancements in hybrid approaches have successfully leveraged the complementary capabilities of Convolutional Neural Networks (ConvNets) and Vision Transformers (ViTs), surpassing their individual capabilities across various benchmark datasets such as ImageNet. We trained recent SOTA hybrid models from scratch on Oxford-102, CUB-200 and PlantVillage datasets, comparing our proposed hybrid variants against them. Our proposed hybrid variants showed favorable performance in terms of top-1 accuracy compared to other evaluated hybrid models. For instance, our CTRLF-S (+AKF) significantly outperforms PVTv2-B1 by 24.0\% on Oxford-102 dataset with a slight decrease in Params and FLOPs and our CTRLF-S (+CKF) outperforms PVTv2-B1 by 25.7\% on CUB-200 and by 0.16\% on PlantVillage dataset. Meanwhile, our CTRLF-B (+AKF) outperforms PVTv2-B2, FasterViT-0 ,Coat-0, Coat-1 and Moat-0 by 23.8\%, 19.7\%, 12.9\%, 12.2\% and 7.4\% on Oxford-102, with a slight decrease in Params and FLOPS, and our CTRLF-B (+CKF) surpasses PVTv2-B2, FasterViT-0, Coat-0, Coat-1 and Moat-0 by 28.2\%, 22.0\%, 7.5\%, 4.0\% and 22.3\% on CUB-200, and by 0.18\%, 0.15\%, 0.06\%, 0.06\% and 0.07\% on PlantVillage.

\begin{table}[h]
    \centering
    \caption{Comparison with the recent state-of-the-art models belonging to the categories of ConvNets, ViTs and Hybrid models. We train all the models listed in this table from scratch on Oxford-102, CUB-200 and PlantVillage datasets, reporting their respective top-1 accuracy scores, parameters, and FLOPs. For fair comparison, all models' parameters and FLOPs are calculated with an output layer of 102 neurons/classes (as in Oxford-102).}
    \begin{tabular}{lllllll}
        \toprule
        & Model   & Params & FLOPs & Oxford-102  & CUB-200  & PlantVillage  \\
        &   &   &   & top-1 acc & top-1 acc & top-1 acc \\
        \toprule
        \multirow{4}{*}{ConvNets} & MobileNetV3-L \cite{howard_searching_2019}   & 4.3M   & 0.22G & 65.46\% & 39.42\% & 99.79\% \\
                                  & EfficientNet-B3 \cite{tan_efficientnet_2019} & 10.9M  & 1.0G  & 69.85\% & 48.05\% & 99.85\% \\ 
                                  & EfficientNetV2-S \cite{tan_efficientnetv2_2021} & 20.3M  & 2.9G  & 63.36\% & 34.09\% & 99.84\% \\
                                  & ConvNeXt-T \cite{liu_convnet_2022}      & 27.9M  & 4.5G  & 53.40\% & 31.81\% & 99.59\% \\
            
        \cmidrule(lr){1-7}

        \multirow{4}{*}{ViTs}     & ViT-B \cite{dosovitskiy_image_2020}            & 85.9M  & 17.6G & 46.50\% & 12.70\% & 98.74\% \\
                                  & CrossViT-15\dag \cite{chen_crossvit_2021} & 27.7M  & 6.1G  & 48.61\% & 20.75\% & 99.37\% \\ 
                                  & PVTv1-S \cite{wang_pyramid_2021}         & 24.0M  & 3.7G  & 52.85\% & 17.93\% & 99.59\% \\
                                  & Swin-T \cite{liu_swin_2021}          & 27.6M  & 4.5G  & 60.82\% & 28.55\% & 99.71\% \\
        \cmidrule(lr){1-7}

        \multirow{12}{*}{Hybrid}  & PVTv2-B1 \cite{wang_pvt_2022}        & 13.5M  & 2.1G  & 58.40\% & 31.62\% & 99.75\% \\
                                  & PVTv2-B2 \cite{wang_pvt_2022}        & 24.9M  & 3.9G  & 60.22\% & 33.47\% & 99.73\% \\ 
                                  & FasterViT-0 \cite{hatamizadeh_fastervit_2024}     & 30.9M  & 3.3G  & 64.31\% & 39.70\% & 99.76\% \\ 
                                  & Coat-0 \cite{dai_coatnet_2021}          & 25.0M  & 4.2G  & 71.10\% & 54.14\% & 99.85\% \\
                                  & Coat-1 \cite{dai_coatnet_2021}          & 41.0M  & 8.4G  & 71.81\% & 57.66\% & 99.85\% \\
                                  & Moat-0 \cite{yang_moat_2022}          & 27.0M  & 5.7G  & 76.57\% & 39.42\% & 99.84\% \\
                                  & CTRLF-S  (+AKF)      & 9.97M  & 1.43G & 82.42\% & 56.27\% & 99.85\% \\ 
                                  & CTRLF-B  (+AKF)      & 21.36M & 3.29G & 84.01\% & 60.77\% & 99.89\% \\
                                  & CTRLF-S  (+CKF)      & 9.99M  & 1.43G & 81.88\% & 57.35\% & 99.91\% \\
                                  & CTRLF-B  (+CKF)      & 21.39M & 3.29G & 83.00\% & 61.68\% & 99.91\% \\
                                    
        \bottomrule

    \end{tabular}
\end{table}

\section{Conclusion}
In this paper, we introduce CTRL-F, a novel lightweight hybrid architecture that seamlessly integrates the strengths of Convolutional Neural Networks (ConvNets) and Vision Transformers (ViTs) for image classification tasks. To effectively capture global context with reduced complexity, we introduce a Multi-level Feature Cross-Attention Transformer (MFCA) module with a dual-branch structure. This module efficiently operates on low-resolution, multi-level feature maps obtained from two different convolutional stages and process these feature maps independently using a series of transformer encoders and facilitates efficient knowledge exchange between both branches via cross-attention. We also develop two effective strategies for representation learning fusion named: adaptive knowledge fusion (AKF) and collaborative knowledge fusion (CKF), providing complementary benefits by ensuring that the model incorporates both local representations learned from convolution and global representations learned from attention. Extensive experiments demonstrate that our proposed CTRL-F enjoys ConvNets benefits including fast convergence and robust generalization with limited data, alongside the high capacity and global processing capabilities of transformer, achieving state-of-the-art results on different datasets with compact computational cost.

\bibliographystyle{ieeetr}
\bibliography{references}

\begin{thebibliography}{10}

\bibitem{krizhevsky_imagenet_2012}
A.~Krizhevsky, I.~Sutskever, and G.~E. Hinton, ``{ImageNet} {Classification} with {Deep} {Convolutional} {Neural} {Networks},'' in {\em Advances in {Neural} {Information} {Processing} {Systems}}, vol.~25, Curran Associates, Inc., 2012.

\bibitem{simonyan_very_2015}
K.~Simonyan and A.~Zisserman, ``Very deep convolutional networks for large-scale image recognition,'' {\em 3rd International Conference on Learning Representations (ICLR 2015)}, 2015.
\newblock Publisher: Computational and Biological Learning Society.

\bibitem{girshick_rich_2014}
R.~Girshick, J.~Donahue, T.~Darrell, and J.~Malik, ``Rich {Feature} {Hierarchies} for {Accurate} {Object} {Detection} and {Semantic} {Segmentation},'' in {\em 2014 {IEEE} {Conference} on {Computer} {Vision} and {Pattern} {Recognition}}, pp.~580--587, June 2014.
\newblock ISSN: 1063-6919.

\bibitem{long_fully_2015}
J.~Long, E.~Shelhamer, and T.~Darrell, ``Fully convolutional networks for semantic segmentation,'' in {\em 2015 {IEEE} {Conference} on {Computer} {Vision} and {Pattern} {Recognition} ({CVPR})}, pp.~3431--3440, June 2015.
\newblock ISSN: 1063-6919.

\bibitem{vaswani_attention_2017}
A.~Vaswani, N.~Shazeer, N.~Parmar, J.~Uszkoreit, L.~Jones, A.~N. Gomez, L.~Kaiser, and I.~Polosukhin, ``Attention is {All} you {Need},'' in {\em Advances in {Neural} {Information} {Processing} {Systems}}, vol.~30, Curran Associates, Inc., 2017.

\bibitem{wang_non-local_2018}
X.~Wang, R.~Girshick, A.~Gupta, and K.~He, ``Non-local {Neural} {Networks},'' in {\em 2018 {IEEE}/{CVF} {Conference} on {Computer} {Vision} and {Pattern} {Recognition}}, (Salt Lake City, UT, USA), pp.~7794--7803, IEEE, June 2018.

\bibitem{bello_attention_2019}
I.~Bello, B.~Zoph, Q.~Le, A.~Vaswani, and J.~Shlens, ``Attention {Augmented} {Convolutional} {Networks},'' in {\em 2019 {IEEE}/{CVF} {International} {Conference} on {Computer} {Vision} ({ICCV})}, (Seoul, Korea (South)), pp.~3285--3294, IEEE, Oct. 2019.

\bibitem{zhuoran_efficient_2021}
S.~Zhuoran, Z.~Mingyuan, Z.~Haiyu, Y.~Shuai, and L.~Hongsheng, ``Efficient {Attention}: {Attention} with {Linear} {Complexities},'' in {\em 2021 {IEEE} {Winter} {Conference} on {Applications} of {Computer} {Vision} ({WACV})}, (Waikoloa, HI, USA), pp.~3530--3538, IEEE, Jan. 2021.

\bibitem{dosovitskiy_image_2020}
A.~Dosovitskiy, L.~Beyer, A.~Kolesnikov, D.~Weissenborn, X.~Zhai, T.~Unterthiner, M.~Dehghani, M.~Minderer, G.~Heigold, S.~Gelly, J.~Uszkoreit, and N.~Houlsby, ``An {Image} is {Worth} 16x16 {Words}: {Transformers} for {Image} {Recognition} at {Scale},'' Oct. 2020.

\bibitem{deng_imagenet_2009}
J.~Deng, W.~Dong, R.~Socher, L.-J. Li, K.~Li, and L.~Fei-Fei, ``{ImageNet}: {A} large-scale hierarchical image database,'' in {\em 2009 {IEEE} {Conference} on {Computer} {Vision} and {Pattern} {Recognition}}, pp.~248--255, June 2009.
\newblock ISSN: 1063-6919.

\bibitem{sun_revisiting_2017}
C.~Sun, A.~Shrivastava, S.~Singh, and A.~Gupta, ``Revisiting {Unreasonable} {Effectiveness} of {Data} in {Deep} {Learning} {Era},'' in {\em 2017 {IEEE} {International} {Conference} on {Computer} {Vision} ({ICCV})}, pp.~843--852, Oct. 2017.
\newblock ISSN: 2380-7504.

\bibitem{touvron_training_2021}
H.~Touvron, M.~Cord, M.~Douze, F.~Massa, A.~Sablayrolles, and H.~Jegou, ``Training data-efficient image transformers \& distillation through attention,'' in {\em Proceedings of the 38th {International} {Conference} on {Machine} {Learning}}, pp.~10347--10357, PMLR, July 2021.
\newblock ISSN: 2640-3498.

\bibitem{chen_dearkd_2022}
X.~Chen, Q.~Cao, Y.~Zhong, J.~Zhang, S.~Gao, and D.~Tao, ``{DearKD}: {Data}-{Efficient} {Early} {Knowledge} {Distillation} for {Vision} {Transformers},'' in {\em 2022 {IEEE}/{CVF} {Conference} on {Computer} {Vision} and {Pattern} {Recognition} ({CVPR})}, pp.~12042--12052, June 2022.
\newblock ISSN: 2575-7075.

\bibitem{steiner_how_2022}
A.~P. Steiner, A.~Kolesnikov, X.~Zhai, R.~Wightman, J.~Uszkoreit, and L.~Beyer, ``How to train your {ViT}? {Data}, {Augmentation}, and {Regularization} in {Vision} {Transformers},'' {\em Transactions on Machine Learning Research}, Apr. 2022.

\bibitem{yuan_tokens--token_2021}
L.~Yuan, Y.~Chen, T.~Wang, W.~Yu, Y.~Shi, Z.~Jiang, F.~E.~H. Tay, J.~Feng, and S.~Yan, ``Tokens-to-{Token} {ViT}: {Training} {Vision} {Transformers} from {Scratch} on {ImageNet},'' in {\em 2021 {IEEE}/{CVF} {International} {Conference} on {Computer} {Vision} ({ICCV})}, pp.~538--547, Oct. 2021.
\newblock ISSN: 2380-7504.

\bibitem{wang_pyramid_2021}
W.~Wang, E.~Xie, X.~Li, D.-P. Fan, K.~Song, D.~Liang, T.~Lu, P.~Luo, and L.~Shao, ``Pyramid {Vision} {Transformer}: {A} {Versatile} {Backbone} for {Dense} {Prediction} without {Convolutions},'' in {\em 2021 {IEEE}/{CVF} {International} {Conference} on {Computer} {Vision} ({ICCV})}, pp.~548--558, Oct. 2021.
\newblock ISSN: 2380-7504.

\bibitem{chen_crossvit_2021}
C.-F.~R. Chen, Q.~Fan, and R.~Panda, ``{CrossViT}: {Cross}-{Attention} {Multi}-{Scale} {Vision} {Transformer} for {Image} {Classification},'' in {\em 2021 {IEEE}/{CVF} {International} {Conference} on {Computer} {Vision} ({ICCV})}, pp.~347--356, Oct. 2021.
\newblock ISSN: 2380-7504.

\bibitem{liu_swin_2021}
Z.~Liu, Y.~Lin, Y.~Cao, H.~Hu, Y.~Wei, Z.~Zhang, S.~Lin, and B.~Guo, ``Swin {Transformer}: {Hierarchical} {Vision} {Transformer} using {Shifted} {Windows},'' in {\em 2021 {IEEE}/{CVF} {International} {Conference} on {Computer} {Vision} ({ICCV})}, pp.~9992--10002, Oct. 2021.
\newblock ISSN: 2380-7504.

\bibitem{liu_swin_2022}
Z.~Liu, H.~Hu, Y.~Lin, Z.~Yao, Z.~Xie, Y.~Wei, J.~Ning, Y.~Cao, Z.~Zhang, L.~Dong, F.~Wei, and B.~Guo, ``Swin {Transformer} {V2}: {Scaling} {Up} {Capacity} and {Resolution},'' in {\em 2022 {IEEE}/{CVF} {Conference} on {Computer} {Vision} and {Pattern} {Recognition} ({CVPR})}, pp.~11999--12009, June 2022.
\newblock ISSN: 2575-7075.

\bibitem{tan_efficientnetv2_2021}
M.~Tan and Q.~Le, ``{EfficientNetV2}: {Smaller} {Models} and {Faster} {Training},'' in {\em Proceedings of the 38th {International} {Conference} on {Machine} {Learning}}, pp.~10096--10106, PMLR, July 2021.
\newblock ISSN: 2640-3498.

\bibitem{liu_convnet_2022}
Z.~Liu, H.~Mao, C.-Y. Wu, C.~Feichtenhofer, T.~Darrell, and S.~Xie, ``A {ConvNet} for the 2020s,'' in {\em 2022 {IEEE}/{CVF} {Conference} on {Computer} {Vision} and {Pattern} {Recognition} ({CVPR})}, pp.~11966--11976, June 2022.
\newblock ISSN: 2575-7075.

\bibitem{xiao_early_2021}
T.~Xiao, M.~Singh, E.~Mintun, T.~Darrell, P.~Dollar, and R.~Girshick, ``Early {Convolutions} {Help} {Transformers} {See} {Better},'' in {\em Advances in {Neural} {Information} {Processing} {Systems}}, vol.~34, pp.~30392--30400, Curran Associates, Inc., 2021.

\bibitem{xu_vitae_2021}
Y.~Xu, Q.~ZHANG, J.~Zhang, and D.~Tao, ``{ViTAE}: {Vision} {Transformer} {Advanced} by {Exploring} {Intrinsic} {Inductive} {Bias},'' in {\em Advances in {Neural} {Information} {Processing} {Systems}}, vol.~34, pp.~28522--28535, Curran Associates, Inc., 2021.

\bibitem{peng_conformer_2021}
Z.~Peng, W.~Huang, S.~Gu, L.~Xie, Y.~Wang, J.~Jiao, and Q.~Ye, ``Conformer: {Local} {Features} {Coupling} {Global} {Representations} for {Visual} {Recognition},'' in {\em 2021 {IEEE}/{CVF} {International} {Conference} on {Computer} {Vision} ({ICCV})}, (Montreal, QC, Canada), pp.~357--366, IEEE, Oct. 2021.

\bibitem{chen_mobile-former_2022}
Y.~Chen, X.~Dai, D.~Chen, M.~Liu, X.~Dong, L.~Yuan, and Z.~Liu, ``Mobile-{Former}: {Bridging} {MobileNet} and {Transformer},'' in {\em 2022 {IEEE}/{CVF} {Conference} on {Computer} {Vision} and {Pattern} {Recognition} ({CVPR})}, (New Orleans, LA, USA), pp.~5260--5269, IEEE, June 2022.

\bibitem{mehta_mobilevit_2021}
S.~Mehta and M.~Rastegari, ``{MobileViT}: {Light}-weight, {General}-purpose, and {Mobile}-friendly {Vision} {Transformer},'' Oct. 2021.

\bibitem{mehta_separable_2022}
S.~Mehta and M.~Rastegari, ``Separable {Self}-attention for {Mobile} {Vision} {Transformers},'' June 2022.
\newblock arXiv:2206.02680 [cs].

\bibitem{dai_coatnet_2021}
Z.~Dai, H.~Liu, Q.~V. Le, and M.~Tan, ``{CoAtNet}: {Marrying} {Convolution} and {Attention} for {All} {Data} {Sizes},'' in {\em Advances in {Neural} {Information} {Processing} {Systems}}, vol.~34, pp.~3965--3977, Curran Associates, Inc., 2021.

\bibitem{yang_moat_2022}
C.~Yang, S.~Qiao, Q.~Yu, X.~Yuan, Y.~Zhu, A.~Yuille, H.~Adam, and L.-C. Chen, ``{MOAT}: {Alternating} {Mobile} {Convolution} and {Attention} {Brings} {Strong} {Vision} {Models},'' Sept. 2022.

\bibitem{maaz_edgenext_2023}
M.~Maaz, A.~Shaker, H.~Cholakkal, S.~Khan, S.~W. Zamir, R.~M. Anwer, and F.~Shahbaz~Khan, ``{EdgeNeXt}: {Efficiently} {Amalgamated} {CNN}-{Transformer} {Architecture} for {Mobile} {Vision} {Applications},'' in {\em Computer {Vision} – {ECCV} 2022 {Workshops}: {Tel} {Aviv}, {Israel}, {October} 23–27, 2022, {Proceedings}, {Part} {VII}}, (Berlin, Heidelberg), pp.~3--20, Springer-Verlag, Feb. 2023.

\bibitem{cheng_higherhrnet_2020}
B.~Cheng, B.~Xiao, J.~Wang, H.~Shi, T.~S. Huang, and L.~Zhang, ``{HigherHRNet}: {Scale}-{Aware} {Representation} {Learning} for {Bottom}-{Up} {Human} {Pose} {Estimation},'' in {\em 2020 {IEEE}/{CVF} {Conference} on {Computer} {Vision} and {Pattern} {Recognition} ({CVPR})}, pp.~5385--5394, June 2020.
\newblock ISSN: 2575-7075.

\bibitem{chen_big-little_2018}
C.-F.~R. Chen, Q.~Fan, N.~Mallinar, T.~Sercu, and R.~Feris, ``Big-{Little} {Net}: {An} {Efficient} {Multi}-{Scale} {Feature} {Representation} for {Visual} and {Speech} {Recognition},'' Sept. 2018.

\bibitem{yang_pointcat_2023}
X.~Yang, M.~Jin, W.~He, and Q.~Chen, ``{PointCAT}: {Cross}-{Attention} {Transformer} for point cloud,'' Apr. 2023.
\newblock arXiv:2304.03012 [cs].

\bibitem{he_deep_2016}
K.~He, X.~Zhang, S.~Ren, and J.~Sun, ``Deep {Residual} {Learning} for {Image} {Recognition},'' in {\em 2016 {IEEE} {Conference} on {Computer} {Vision} and {Pattern} {Recognition} ({CVPR})}, pp.~770--778, June 2016.
\newblock ISSN: 1063-6919.

\bibitem{szegedy_going_2015}
C.~Szegedy, W.~Liu, Y.~Jia, P.~Sermanet, S.~Reed, D.~Anguelov, D.~Erhan, V.~Vanhoucke, and A.~Rabinovich, ``Going deeper with convolutions,'' in {\em 2015 {IEEE} {Conference} on {Computer} {Vision} and {Pattern} {Recognition} ({CVPR})}, pp.~1--9, June 2015.
\newblock ISSN: 1063-6919.

\bibitem{chollet_xception_2017}
F.~Chollet, ``Xception: {Deep} {Learning} with {Depthwise} {Separable} {Convolutions},'' in {\em 2017 {IEEE} {Conference} on {Computer} {Vision} and {Pattern} {Recognition} ({CVPR})}, pp.~1800--1807, July 2017.
\newblock ISSN: 1063-6919.

\bibitem{huang_densely_2017}
G.~Huang, Z.~Liu, L.~Van Der~Maaten, and K.~Q. Weinberger, ``Densely {Connected} {Convolutional} {Networks},'' in {\em 2017 {IEEE} {Conference} on {Computer} {Vision} and {Pattern} {Recognition} ({CVPR})}, pp.~2261--2269, July 2017.
\newblock ISSN: 1063-6919.

\bibitem{howard_mobilenets_2017}
A.~G. Howard, M.~Zhu, B.~Chen, D.~Kalenichenko, W.~Wang, T.~Weyand, M.~Andreetto, and H.~Adam, ``{MobileNets}: {Efficient} {Convolutional} {Neural} {Networks} for {Mobile} {Vision} {Applications},'' Apr. 2017.

\bibitem{sandler_mobilenetv2_2018}
M.~Sandler, A.~Howard, M.~Zhu, A.~Zhmoginov, and L.-C. Chen, ``{MobileNetV2}: {Inverted} {Residuals} and {Linear} {Bottlenecks},'' in {\em 2018 {IEEE}/{CVF} {Conference} on {Computer} {Vision} and {Pattern} {Recognition}}, pp.~4510--4520, June 2018.
\newblock ISSN: 2575-7075.

\bibitem{howard_searching_2019}
A.~Howard, M.~Sandler, B.~Chen, W.~Wang, L.-C. Chen, M.~Tan, G.~Chu, V.~Vasudevan, Y.~Zhu, R.~Pang, H.~Adam, and Q.~Le, ``Searching for {MobileNetV3},'' in {\em 2019 {IEEE}/{CVF} {International} {Conference} on {Computer} {Vision} ({ICCV})}, pp.~1314--1324, Oct. 2019.
\newblock ISSN: 2380-7504.

\bibitem{zhang_shufflenet_2018}
X.~Zhang, X.~Zhou, M.~Lin, and J.~Sun, ``{ShuffleNet}: {An} {Extremely} {Efficient} {Convolutional} {Neural} {Network} for {Mobile} {Devices},'' in {\em 2018 {IEEE}/{CVF} {Conference} on {Computer} {Vision} and {Pattern} {Recognition}}, pp.~6848--6856, June 2018.
\newblock ISSN: 2575-7075.

\bibitem{ma_shufflenet_2018}
N.~Ma, X.~Zhang, H.-T. Zheng, and J.~Sun, ``{ShuffleNet} {V2}: {Practical} {Guidelines} for {Efficient} {CNN} {Architecture} {Design},'' in {\em Computer {Vision} – {ECCV} 2018} (V.~Ferrari, M.~Hebert, C.~Sminchisescu, and Y.~Weiss, eds.), (Cham), pp.~122--138, Springer International Publishing, 2018.

\bibitem{tan_efficientnet_2019}
M.~Tan and Q.~Le, ``{EfficientNet}: {Rethinking} {Model} {Scaling} for {Convolutional} {Neural} {Networks},'' in {\em Proceedings of the 36th {International} {Conference} on {Machine} {Learning}}, pp.~6105--6114, PMLR, May 2019.
\newblock ISSN: 2640-3498.

\bibitem{hu_squeeze-and-excitation_2018}
J.~Hu, L.~Shen, and G.~Sun, ``Squeeze-and-{Excitation} {Networks},'' in {\em 2018 {IEEE}/{CVF} {Conference} on {Computer} {Vision} and {Pattern} {Recognition}}, pp.~7132--7141, June 2018.
\newblock ISSN: 2575-7075.

\bibitem{zhang_mixup_2018}
H.~Zhang, M.~Cisse, Y.~N. Dauphin, and D.~Lopez-Paz, ``mixup: {Beyond} {Empirical} {Risk} {Minimization},'' Feb. 2018.

\bibitem{cubuk_randaugment_2020}
E.~D. Cubuk, B.~Zoph, J.~Shlens, and Q.~V. Le, ``Randaugment: {Practical} automated data augmentation with a reduced search space,'' in {\em 2020 {IEEE}/{CVF} {Conference} on {Computer} {Vision} and {Pattern} {Recognition} {Workshops} ({CVPRW})}, (Seattle, WA, USA), pp.~3008--3017, IEEE, June 2020.

\bibitem{yun_cutmix_2019}
S.~Yun, D.~Han, S.~Chun, S.~J. Oh, Y.~Yoo, and J.~Choe, ``{CutMix}: {Regularization} {Strategy} to {Train} {Strong} {Classifiers} {With} {Localizable} {Features},'' in {\em 2019 {IEEE}/{CVF} {International} {Conference} on {Computer} {Vision} ({ICCV})}, (Seoul, Korea (South)), pp.~6022--6031, IEEE, Oct. 2019.

\bibitem{el-assiouti_regioninpaint_2023}
O.~S. El-Assiouti, G.~Hamed, H.~El-Saadawy, H.~M. Ebied, and D.~Khattab, ``{RegionInpaint}, {Cutoff} and {RegionMix}: {Introducing} {Novel} {Augmentation} {Techniques} for {Enhancing} the {Generalization} of {Brain} {Tumor} {Identification},'' {\em IEEE Access}, vol.~11, pp.~83232--83250, 2023.
\newblock Conference Name: IEEE Access.

\bibitem{radosavovic_designing_2020}
I.~Radosavovic, R.~P. Kosaraju, R.~Girshick, K.~He, and P.~Dollar, ``Designing {Network} {Design} {Spaces},'' in {\em 2020 {IEEE}/{CVF} {Conference} on {Computer} {Vision} and {Pattern} {Recognition} ({CVPR})}, (Seattle, WA, USA), pp.~10425--10433, IEEE, June 2020.

\bibitem{krizhevsky_learning_2012}
A.~Krizhevsky, ``Learning {Multiple} {Layers} of {Features} from {Tiny} {Images},'' {\em University of Toronto}, May 2012.

\bibitem{nilsback_automated_2008}
M.-E. Nilsback and A.~Zisserman, ``Automated {Flower} {Classification} over a {Large} {Number} of {Classes},'' in {\em 2008 {Sixth} {Indian} {Conference} on {Computer} {Vision}, {Graphics} \& {Image} {Processing}}, pp.~722--729, Dec. 2008.

\bibitem{zhao_cumulative_2023}
B.~Zhao, R.~Song, and J.~Liang, ``Cumulative {Spatial} {Knowledge} {Distillation} for {Vision} {Transformers},'' in {\em 2023 {IEEE}/{CVF} {International} {Conference} on {Computer} {Vision} ({ICCV})}, (Paris, France), pp.~6123--6132, IEEE, Oct. 2023.

\bibitem{wu_cvt_2021}
H.~Wu, B.~Xiao, N.~Codella, M.~Liu, X.~Dai, L.~Yuan, and L.~Zhang, ``{CvT}: {Introducing} {Convolutions} to {Vision} {Transformers},'' in {\em 2021 {IEEE}/{CVF} {International} {Conference} on {Computer} {Vision} ({ICCV})}, pp.~22--31, Oct. 2021.
\newblock ISSN: 2380-7504.

\bibitem{graham_levit_2021}
B.~Graham, A.~El-Nouby, H.~Touvron, P.~Stock, A.~Joulin, H.~Jégou, and M.~Douze, ``{LeViT}: a {Vision} {Transformer} in {ConvNet}’s {Clothing} for {Faster} {Inference},'' in {\em 2021 {IEEE}/{CVF} {International} {Conference} on {Computer} {Vision} ({ICCV})}, pp.~12239--12249, Oct. 2021.
\newblock ISSN: 2380-7504.

\bibitem{wang_pvt_2022}
W.~Wang, E.~Xie, X.~Li, D.-P. Fan, K.~Song, D.~Liang, T.~Lu, P.~Luo, and L.~Shao, ``{PVT} v2: {Improved} baselines with {Pyramid} {Vision} {Transformer},'' {\em Computational Visual Media}, vol.~8, pp.~415--424, Sept. 2022.

\bibitem{hatamizadeh_fastervit_2024}
A.~Hatamizadeh, G.~Heinrich, H.~Yin, A.~Tao, J.~M. Alvarez, J.~Kautz, and P.~Molchanov, ``{FASTERVIT}: {FAST} {VISION} {TRANSFORMERS} {WITH} {HIERARCHICAL} {ATTENTION},'' 2024.

\bibitem{hendrycks_gaussian_2023}
D.~Hendrycks and K.~Gimpel, ``Gaussian {Error} {Linear} {Units} ({GELUs}),'' June 2023.
\newblock arXiv:1606.08415 [cs].

\bibitem{srivastava_dropout_2014}
N.~Srivastava, G.~Hinton, A.~Krizhevsky, I.~Sutskever, and R.~Salakhutdinov, ``Dropout: {A} {Simple} {Way} to {Prevent} {Neural} {Networks} from {Overfitting},'' {\em Journal of Machine Learning Research}, vol.~15, no.~56, pp.~1929--1958, 2014.

\bibitem{mohanty_using_2016}
S.~P. Mohanty, D.~P. Hughes, and M.~Salathé, ``Using {Deep} {Learning} for {Image}-{Based} {Plant} {Disease} {Detection},'' {\em Frontiers in Plant Science}, vol.~7, Sept. 2016.
\newblock Publisher: Frontiers.

\bibitem{WahCUB_200_2011}
C.~Wah, S.~Branson, P.~Welinder, P.~Perona, and S.~Belongie, ``Caltech-ucsd birds-200-2011 (cub-200-2011),'' Tech. Rep. CNS-TR-2011-001, California Institute of Technology, 2011.

\bibitem{loshchilov_decoupled_2018}
I.~Loshchilov and F.~Hutter, ``Decoupled {Weight} {Decay} {Regularization},'' Sept. 2018.

\end{thebibliography}

\end{document}